\documentclass{article}

\usepackage{microtype}
\usepackage{graphicx}
\usepackage{subfigure}
\usepackage{booktabs} 
\usepackage{url}
\usepackage{breakurl}

\usepackage{amsmath}
\usepackage{multirow}

\usepackage[accepted]{sysml2019}

\sysmltitlerunning{Submission and Formatting Instructions for SysML 2019}

\begin{document}

\twocolumn[
\sysmltitle{Compressing Language Models using Doped Kronecker Products}



\sysmlsetsymbol{equal}{*}

\begin{sysmlauthorlist}
\sysmlauthor{Urmish Thakker}{arm}
\sysmlauthor{Paul Whatmough}{arm}
\sysmlauthor{Zhi-Gang Liu}{arm}
\sysmlauthor{Matthew Mattina}{arm}
\sysmlauthor{Jesse Beu}{arm}
\end{sysmlauthorlist}

\sysmlaffiliation{arm}{Arm ML Research Lab}

\sysmlcorrespondingauthor{Urmish Thakker}{urmish.thakker@arm.com}

\sysmlkeywords{Machine Learning, SysML}

\vskip 0.3in
\begin{abstract}
Kronecker Products (KP) have been used to compress IoT RNN Applications by 15-38x compression factors, achieving better results than traditional compression methods. However when KP is applied to large Natural Language Processing tasks, it leads to significant accuracy loss (approx 26\%). This paper proposes a way to recover accuracy otherwise lost when applying KP to large NLP tasks, by allowing additional degrees of freedom in the KP matrix. More formally, we propose doping, a process of adding an extremely sparse overlay matrix on top of the pre-defined KP structure. We call this compression method \textit{doped kronecker product compression}. To train these models,  we present a new solution to the phenomenon of co-matrix adaption (CMA), which uses a new regularization scheme called co-matrix dropout regularization (CMR). 
We present experimental results that demonstrate compression of a large language model with LSTM layers of size 25 MB by 25$\times$ with 1.4\% loss in perplexity score. At 25$\times$ compression, an equivalent pruned network leads to 7.9\% loss in perplexity score, while HMD and LMF lead to 15\% and 27\% loss in perplexity score respectively.
\end{abstract}

]



\printAffiliationsAndNotice{}  
\section{Introduction}
The large size of Natural Language Processing (NLP) applications can make it impossible for them to run on resource constrained devices with limited memory and cache budgets \cite{urmtha01RNN,skiprnn}. Fitting these applications into IoT devices requires significant compression. For example, to fit a 25 MB Language Model on an IoT device with 1 MB L2 Cache, requires 25x compression or 96\% reduction in the number of parameters. Recently, Kronecker Products (KP) were used to compress IoT applications by 15-38x compression factors \cite{urmKron1,urmkron2} and achieves better accuracy than pruning \cite{suyog}, low-rank matrix factorization (LMF) and small baseline (SB). However, when we apply KP to a large language modeling (LM) application, we see a 26\% loss in accuracy at 338x compression. Unlike pruning (amount of sparsity) and LMF (rank of the matrix), there is no obvious method to control the amount of compression of the KP compressed network. \cite{urmkron2} propose Hybrid KP (HKP) to solve this issue. HKP helps recover the lost accuracy by injecting more parameters in the KP compressed network. However, the compression factor reduces to 5x to bring down the accuracy loss within 1.5\% loss in baseline perplexity.

\begin{figure}[tb]
\vskip 0.1in
\centering
\includegraphics[width=\columnwidth]{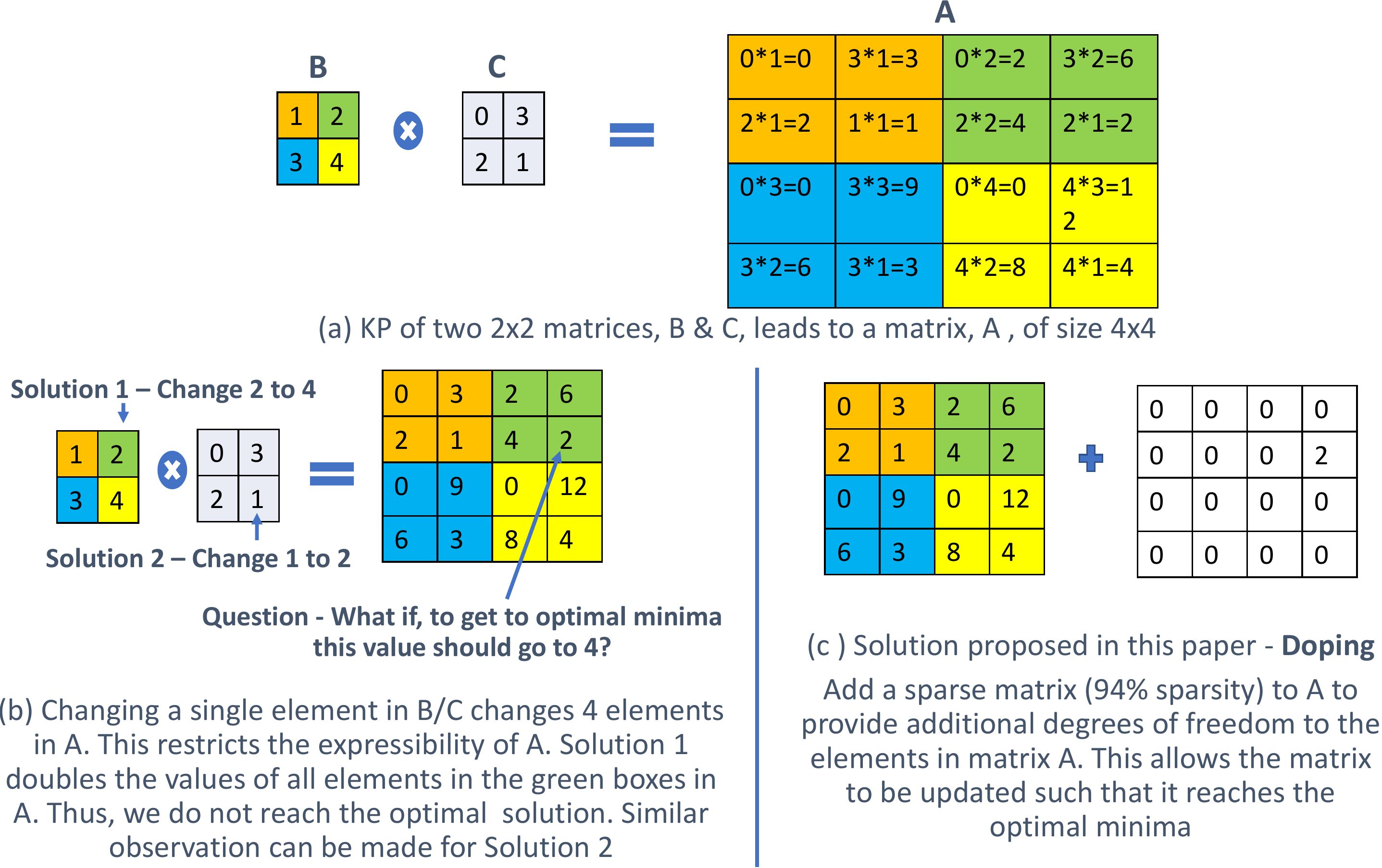}
\caption{(a) Example of a Kronecker Product of two matrices. (b) Issues with back-propagation through a matrix expressed as a KP of two smaller matrices. (c) Shows how doping solves the issues discussed in (b)}
\label{dkp-fig}
\vskip -0.1in
\end{figure}

This paper explores another method to inject parameters into a KP compressed network. This method is based on the observations that parameters in the KP space need additional degrees of freedom (Figure \ref{dkp-fig} a,b). Inspired by robust PCA techniques, we propose adding a sparse matrix to a KP compressed matrix in order to facilitate these additional degrees of freedom (Figure \ref{dkp-fig} c). Thus, a parameter matrix in an RNN, LSTM, GRU, or Transformer layer is replaced by a sum of two matrices -- one expressed as a KP of two smaller matrices ($M_{kp}$) and the other an extremely sparse matrix ($M_{sp}$). During training, $M_{sp}$ starts off with 0\% sparsity. Over time, we prune the unimportant weights in the $M_{sp}$ matrix to get to the required amount of sparsity. These pruned values will represent the equivalent values in $M_{kp}$ that did not require the additional degrees of freedom. This methodology of compression is called Doped Kronecker Product (DKP) in this paper. However, training DKP compressed networks is non-trivial and requires overcoming co-matrix adaption (CMA) (Section \ref{sec-cma}) using a specialized regularization scheme (Section \ref{sec-cmr}).The preliminary results using this compression scheme are encouraging. We show that we can compress the medium sized LM in \cite{ptblm} by 25$\times$ with 1.2\% loss in perplexity score, improving perplexity of pruned \cite{suyog} network by to 6.7\%, HMD \cite{thakker2019runtime} by 13.8\% and LMF \cite{DBLP:journals/corr/KuchaievG17} by 25.8\%.

In the rest of the paper, we discuss the CMA issues associated with a general doping mechanism (\ref{sec-cma}), some methods to overcome CMA based on popular training techniques (\ref{sec-overcma}), the technique proposed in this paper to overcome CMA (\ref{sec-cmr}), results of compressing a medium LM using DKP and comparison against popular compression techniques and previously published work (\ref{sec-results})
\section{Related Work}
The research in NN compression can be broadly categorized under 4 topics - Pruning \cite{han2015deep_compression,suyog}, structured matrix based techniques \cite{circular2,urmkron2,urmKron1,GopeCVPR2020}, quantization \cite{Quant-hubara,Quant-bengio,Quant6,GopeMLSys2019} and tensor decomposition \cite{tjandra2017compressing}. DKP combines pruning and structured matrix based techniques and compares the results with pruning, structured matrix and tensor decomposition based compression techniques. The networks compressed using this technique can be further compressed using quantization.

\section{Doped Kronecker Product (DKP) Compression}
\begin{table}[htb]
\centering
\caption{Results of compressing using DKP when a matrix is expressed as shown in equation \ref{eq:dkp}}
\label{tab:dkp1}
\begin{tabular}{|c|c|c|}
\hline
Baseline Perplexity & \multicolumn{2}{c|}{82.04} \\ \hline
Compression Factor & $338\times$ & $100\times$ \\ \hline
Sparsity of $M_{sp}$ & 100\% & 99.93\% \\ \hline
DKP Perplexity & 104 & 138.3 \\ \hline
\end{tabular}
\end{table}

DKP expresses a matrix as a sum of a $M_{kp}$ and a sparse matrix -
\begin{equation}
    W = B\otimes C + M_{sp}
    \label{eq:dkp}
\end{equation}
The sparsity of $M_{sp}$ determines the amount of compression. For example, if W is of size 100$\times$100, B and C are of size 10$\times$10, then 95\% sparsity in $M_{sp}$ will lead to 14$\times$ compression and 90\% sparsity in $M_{sp}$ will lead to 8.4$\times$ compression. During the initial phase of training, $M_{sp}$ is dense. As training progresses, $M_{sp}$ reaches the required sparsity level. Thus we allow back-propagation to determine which elements of the $M_{kp}$ matrix ($B\otimes C$) require additional degrees of freedom. 

\subsection{Co-matrix Adaptation (CMA)}
\label{sec-cma}

\begin{figure}[tb]
\vskip 0.1in
\centering
\includegraphics[width=\columnwidth]{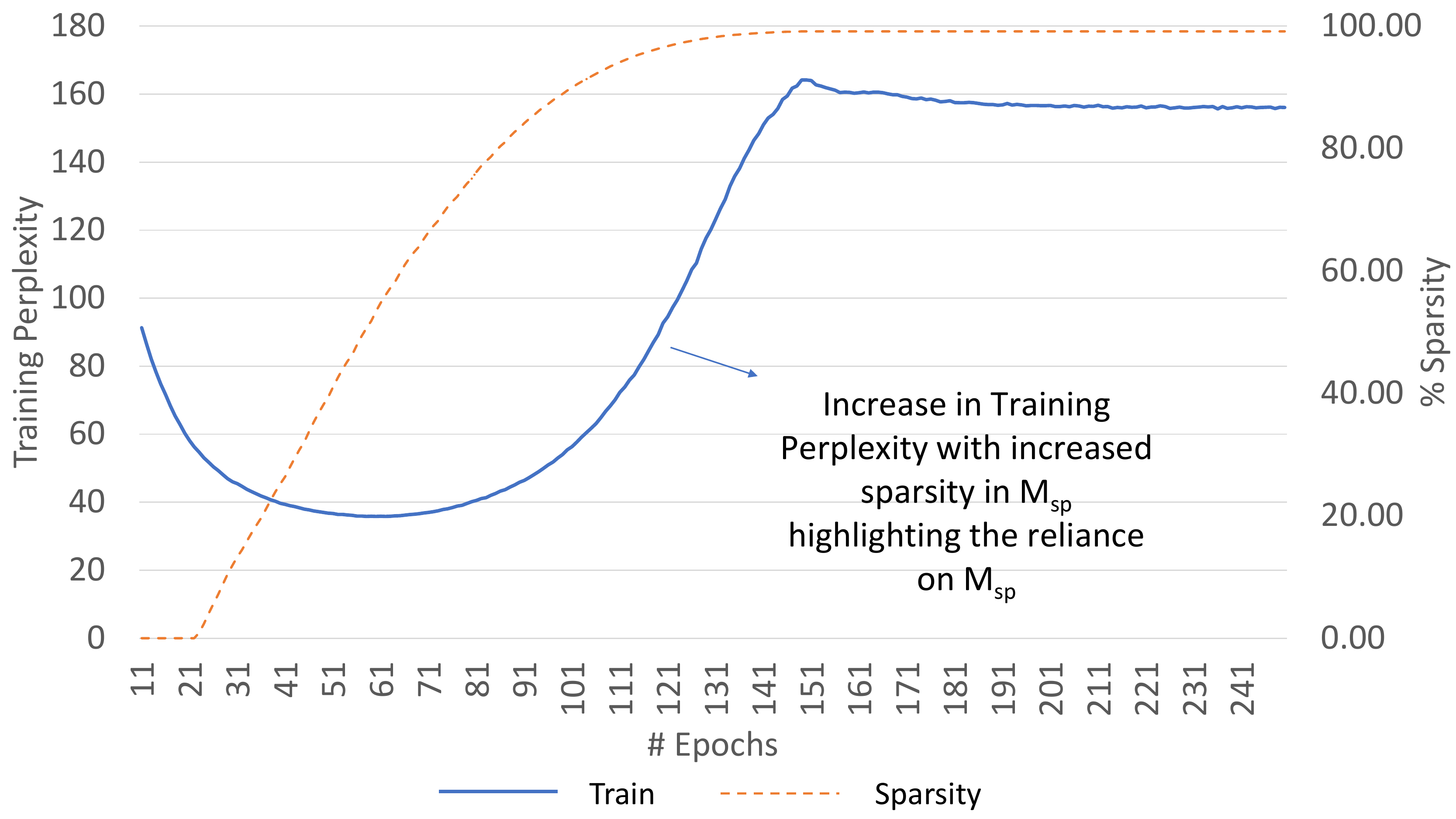}
\caption{Graph of training perplexity vs training epochs and sparsity of $M_{sp}$ matrix vs training epoch for the medium LM at 100$\times$ compression factor. As the sparsity of $M_{sp}$ matrix increases, the training perplexity degrades. This indicates that the NN is too reliant on the $M_{sp}$ matrix during the initial phase of the training process and less reliant on the $M_{kp}$ matrix. We call this phenomenon as co-matrix adaptation.}
\label{cma-fig}
\vskip -0.1in
\end{figure}

Equation \ref{eq:dkp} is one way to implement DKP and represents our initial attempt at compressing using DKP. We compressed the LSTM layers in the medium LM in \cite{ptblm} using this method. The LM has 2 LSTM layers with hidden vector of size 650. This creates matrices of size $2600\times 1300$ amounting to a total size of 25 MB. We compress these layers by 25$\times$ by replacing the matrices in the LSTM layers as shown in equation \ref{eq:dkp}. By adding a $M_{sp}$ with 99\% sparsity, the perplexity score degrades by 32.9\%. Thus adding 1\% more parameters to $M_{kp}$ leads to poorer perplexity score than baseline. Figure \ref{cma-fig} shows the graph of training perplexity vs training epochs and sparsity of $M_{sp}$ matrix vs training epoch for the medium LM at 100$\times$ compression factor. As the sparsity of $M_{sp}$ matrix increases, the training perplexity degrades. This indicates that the LM is too reliant on the $M_{sp}$ matrix during the initial phase of the training process when the matrix is dense and less reliant on the $M_{kp}$ matrix. When the $M_{sp}$ becomes extremely sparse, $M_{kp}$ is no longer able to pull the perplexity score back. We suspect that this might be because the model is stuck in a local minima dictated by the dense $M_{sp}$ matrix. We will refer to this phenomena as co-matrix adaptation (CMA).

The phenomenon of CMA is more clearer when we focus on the number of back-propagation updates during the initial phase of training. The $M_{kp}$ matrix is composed of the kronecker product of two matrices of size $13\times 65$ and $50\times 20$ leading to a total of 9980 parameters. While the $M_{sp}$ matrix in its initial dense form has a total of 3382600 parameters. Thus, during back-propagation, $M_{sp}$ matrix receives $339\times$ more updates than the $M_{kp}$ matrix. This might sway the NN to find a minima that is too reliant on the parameters of the $M_{sp}$ matrix. As a result, when the training progresses and the $M_{sp}$ matrix is pruned, the accuracy drops significantly.

\subsection{Overcoming CMA}
\label{sec-overcma}
\begin{figure}[tb]
\vskip 0.1in
\centering
\includegraphics[width=\columnwidth]{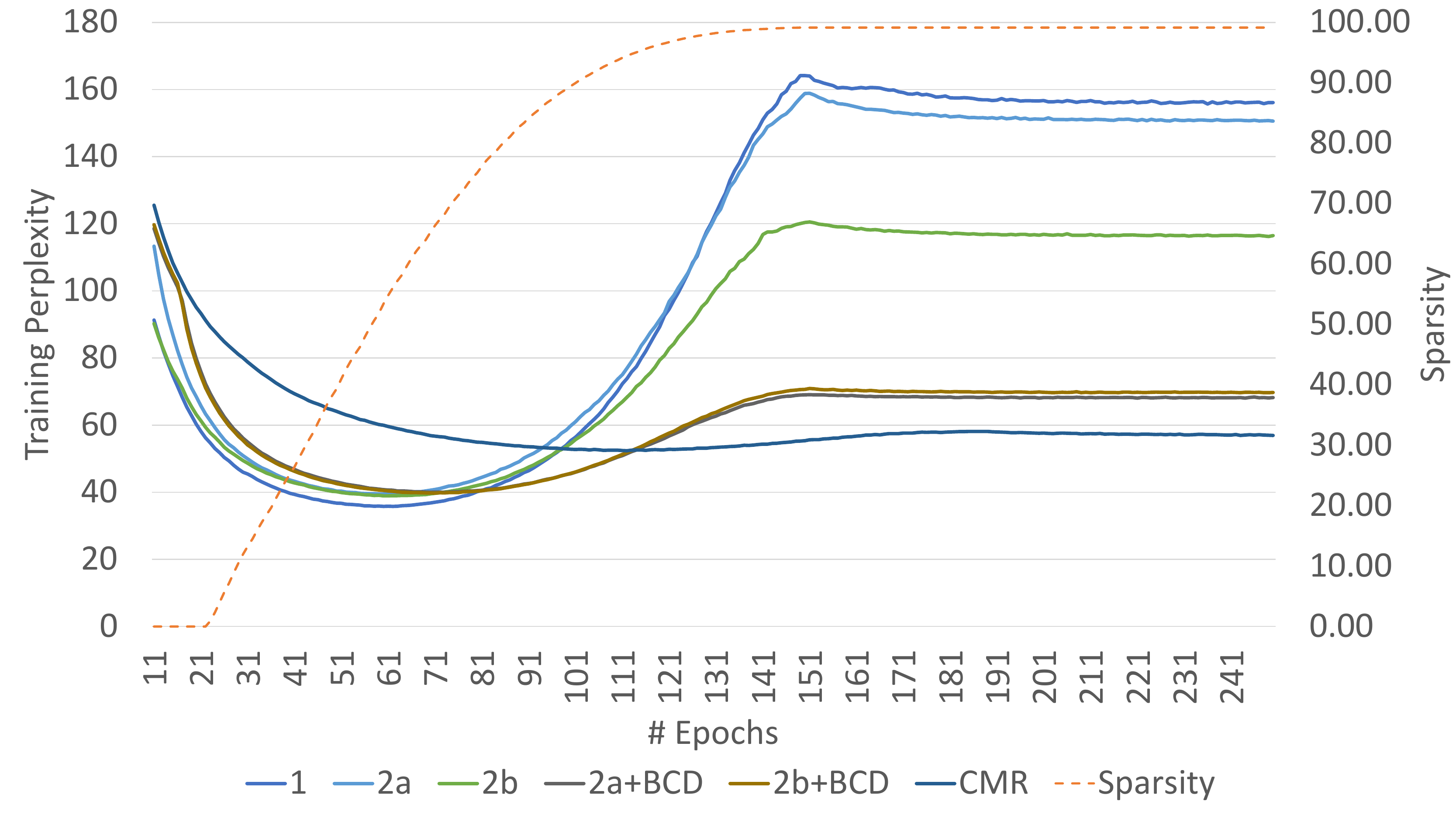}
\caption{Using the various techniques described in equation \ref{eq:dkpcma120} - \ref{eq:dkpcma3456} with and without BCD, we see that the reliance on $M_{sp}$ has reduced. The increase in training perplexity that was visible in \ref{cma-fig} has been managed considerably.}
\label{overcomecma-fig}
\vskip -0.1in
\end{figure}

\begin{table}[htp]
\centering
\caption{Test Perplexity of DKP compressed medium LM for various training techniques discussed in Section \ref{sec-overcma} and \ref{sec-cmr}. CMR techniques leads to the best accuracy for 100$\times$ compression factor.}
\label{tab:cma-test}
\begin{tabular}{|l|l|l|l|}
\hline
\begin{tabular}[c]{@{}l@{}}Compression\\   Factor\end{tabular} & \begin{tabular}[c]{@{}l@{}}Training \\ Method\end{tabular} & \begin{tabular}[c]{@{}l@{}}Sparsity \\ of $M_{sp}$\end{tabular} & \begin{tabular}[c]{@{}l@{}}Test \\ Perplexity\end{tabular} \\ \hline
1x & Baseline & NA & 82.04 \\ \hline
$338\times$ & 2a & 0 & 104.061 \\ \hline
\multirow{7}{*}{$100\times$} & 1 & 99.93 & 150.737 \\ \cline{2-4} 
 & 2a & 99.93 & 138.31 \\ \cline{2-4} 
 & 2b & 99.93 & 123.835 \\ \cline{2-4} 
 & 2a+BCD & 99.93 & 100.37 \\ \cline{2-4} 
 & 2b+BCD & 99.93 & 101.987 \\ \cline{2-4} 
 & CMR & 99.93 & 95.382 \\ \hline
\end{tabular}
\end{table}

\begin{table*}[htb]
\centering
\caption{Results of compressing medium LM over multiple compression factors using DKP, HMD, HKP, LMF, Pruning and Small Baseline}
\label{tab:results}
\begin{tabular}{|c|c|c|c|c|c|c|c|c|c|}
\hline
\begin{tabular}[c]{@{}c@{}}Baseline \\ Test Perplexity\end{tabular} & \multicolumn{9}{c|}{82.04} \\ \hline
\begin{tabular}[c]{@{}c@{}}Compression\\   Factor\end{tabular} & $338\times$ & $100\times$ & $92\times$& $75\times$ & $50\times$ & $25\times$ & $20\times$ & $10\times$ & $5\times$ \\ \hline
DopedKP & 104.061 & 95.49 & 86.576 & 86.73 & 85.45 & 83.24 & 82.94 & 82.9 & 82.53 \\ \hline
Prune & 115.62 & 103.219 & 103.34 & 91.618 & 90.314 & 88.555 & 85.14 & 82.551 & 82.47 \\ \hline
HKD & \multicolumn{6}{c|}{\multirow{4}{*}{Did not run}} & 99.882 & 95.12 & 92.56 \\ \cline{1-1} \cline{8-10} 
HMD & \multicolumn{6}{c|}{} & 105.43 & 97.59 & 95.387 \\ \cline{1-1} \cline{8-10} 
LMF & \multicolumn{6}{c|}{} & 108.61 & 103.42 & 99.29 \\ \cline{1-1} \cline{8-10} 
Small Baseline & \multicolumn{6}{c|}{} & 115.34 & 109.78 & 102.2 \\ \hline
\end{tabular}
\end{table*}

The key to overcoming CMA is to reduce the reliance on $M_{sp}$ initially. This paper explored multiple avenues to do so. 
\begin{subequations}
\label{eq:dkpcma}

\begin{align}
        W&=B \otimes C + \beta \times M_{sp},  min \left\lVert \beta \right\rVert      
        \label{eq:dkpcma120}
\end{align}
\begin{align}
        \begin{split}
            W =\alpha \times (B \otimes C) + \beta \times M_{sp}, \\
            min (\left\lVert \beta \right\rVert + \left\lVert 1/\alpha \right\rVert)        \label{eq:dkpcma3456}
        \end{split}        
\end{align}
\end{subequations}

Each of the equations \ref{eq:dkpcma120} - \ref{eq:dkpcma3456} can be further trained with or without Block Coordinate Descent (BCD). In BCD we alternate between, only training $M_{kp}$, blocking gradient flow to $M_{sp}$, or train $M_{kp}$, blocking gradient flow to $M_{sp}$. The training curves across multiple epochs for these various techniques can be found in Figure \ref{overcomecma-fig}. As the sparsity increases, the training perplexity does not increase as much as in figure \ref{cma-fig} for equations \ref{eq:dkpcma120}-\ref{eq:dkpcma3456} when trained using BCD. However, there is still a small increase in training error with increased sparsity. This can indicate that CMA may have not been completely managed. Table \ref{tab:cma-test} shows the test perplexity at the end of training for the various techniques discussed above. As you can see, these techniques help us bring the perplexity down to 100.37 from 150.737 originally. However, by reducing the compression factor from $338\times$ to $100\times$, we are improving the perplexity by approximately $4$ points only.

\subsection{Co-matrix Row Dropout Regularization (CMR)}
\label{sec-cmr}
To better manage CMA, we focused on how a DKP Cell converts input feature vector into an output feature vector. When an input feature vector, $i$, is fed to a LSTM layer, it gets multiplied with the weight matrix,
\begin{equation}
    o = W*i.
\end{equation}
In the case of DKP, $W$ is composed of two sets of matrices
\begin{equation}
    o = (M_{kp}+M_{sp})*i, \mbox{where }M_{kp} = B\otimes C \\
\end{equation}
\begin{equation}
    o = M_{kp}*i + M_{sp}*i
\end{equation}
Thus each element of the output vector is a combination of output of $M_{kp}*I$ and $M_{sp}*I$, i.e.
\begin{equation}
    o_{j:} = (M_{kp})_{j:}*i + (M_{sp})_{j:}*i
    \label{eq:cmaoutput}
\end{equation}
where $j:$ refers to the $j^{th}$ row of the $M_{kp}$ and $M_{sp}$ matrix. Thus each element (or neuron) of the output feature vector is the sum of elements (or neurons) coming in from the $M_{kp}$ matrix and the $M_{sp}$ matrix.

Our hypothesis is that during CMA, the incoming neurons from the $M_{kp}$ matrix and the $M_{sp}$ matrix learn to co-adapt, leading to lost capacity. Furthermore, because of the dominance of the $M_{sp}$ matrix during the initial phase of the training (Section \ref{sec-cma}), the $M_{kp}$ neurons rely on the $M_{sp}$ neurons heavily. If we introduce a stochastic behavior where either the $M_{kp}$ neuron or the $M_{sp}$ neuron are not available to drive the output neuron, this co-adaptation could be managed. Thus to manage CMA more efficiently, this paper proposes co-matrix row dropout regularization (CMR). This regularization extends the concept of stochastic depth \cite{stochastic} to regularize the output of each row of the output vector in order to avoid CMA. From a mathematical point of view, we introduce dropout after the output of each $M_{kp}*i$ and $M_{sp}*$ value, i.e. equation \ref{eq:cmaoutput} is changed to,

\begin{equation}
    o_{j:} = ((M_{kp})_{j:}*I)bern_1 + ((M_{sp})_{j:}*I)*bern_2,
    \label{eq:cmr}
\end{equation}
where, 
\begin{equation}
    bern_1,bern_2 \sim Bernoulli\{p\}. 
\end{equation}

As the sparsity of $M_{sp}$ increases, the need for CMR decreases and can be removed entirely. The training methodology described by equation \ref{eq:cmr} is referred to as CMR in this paper. CMR is an extremely effective technique to manage CMA as evident by the trends in Figure \ref{overcomecma-fig} for CMR. The training perplexity during the training phase does not increase as the sparsity of the $M_{sp}$ matrix increases. The benefits in the final Test Perplexity are also evident as shown in the last row of the Table \ref{tab:cma-test}.

\section{Results}
\label{sec-results}
\begin{table}[]
\centering
\caption{Comparing DKP with previous published work targeting the same benchmark}
\label{tab:compare-papers}
\begin{tabular}{|l|l|l|}
\hline
\begin{tabular}[c]{@{}l@{}}Comparisons\\   with prior art\end{tabular} & \begin{tabular}[c]{@{}l@{}}Compression\\ Factor\end{tabular} & Test Perp \\ \hline
Baseline LM & $1\times$ & 82.04 \\ \hline
4-bit quant \cite{2017_c1} & $8\times$& 83.84 \\ \hline
3-bit quant \cite{2018_c2} & $10.67\times$& 83.14 \\ \hline
Tensor Train \cite{2019_c3} & $1.67\times$& 168.639 \\ \hline
Weight Distortion \cite{2019_c4} & $10\times$ & 84.64 \\ \hline
Weight Distortion \cite{2019_c4} & $20\times$ & 93.39 \\ \hline
DKP (Ours) & $25\times$ & 83.24 \\ \hline
\end{tabular}
\end{table}

We compress the PTB based medium LM in \cite{ptblm} by multiple compression factors and compare the DKP trained using CMR with pruning (\cite{suyog}), LMF (\cite{DBLP:journals/corr/KuchaievG17}), HMD (\cite{urmtha01RNN}) and HKP (\cite{urmkron2}). As a baseline, we also train a small baseline by reducing the size of the hidden vector in the LSTM layer. 

Table \ref{tab:results} shows the results of compressing the benchmark for various compression factors. As shown, DKP outperforms all compression techniques up to $20\times$ compression factors. Table \ref{tab:compare-papers} further compares these results with other recently published work. Again, our compression technique outperforms these recent papers, achieving $2.5\times$ more compression than the best performing technique.

\section{Conclusion}
This paper presents a new compression technique called Doped Kronecker Product (DKP).
However, training DKP is non-trivial and can run into co-matrix adaptation issues. We further propose co-matrix row dropout regularization (CMR) to manage CMA. The preliminary results demonstrate that using DKP with CMR, we can compress a large language model with LSTM layers of size 25 MB by 25$\times$, with 1.2\% loss in perplexity score. Our technique outperforms popular compression techniques in previously published work, improving the perplexity scores by 7.9\% - 27\%.
\bibliographystyle{sysml2019}
\bibliography{Main}

\end{document}